# Creutzfeldt-Jakob Disease Prediction Using Machine Learning Techniques

Arnav Bhakta[1] and Carolyn Byrne[2]

[1]Phillips Academy Andover
[2] Chicago College of Osteopathic Medicine
at Midwestern University

## Abstract

**Creutzfeldt-Jakob disease (CJD) is a rapidly progressive and fatal neurodegenerative disease, that causes approximately 350 deaths in the United States every year. In specific, it is a prion disease that is caused by a misfolded prion protein, termed PrP$^{Sc}$, which is the infectious form of the prion protein PrP$^{C}$. Rather than being recycled by the body, the PrP$^{Sc}$ aggregates in the brain as plaques, leading to neurodegeneration of surrounding cells and the spongiform characteristics of the pathology. However, there has been very little research done into factors that can affect one's chances of acquiring PrP$^{Sc}$. In this paper, Elastic Net Regression, Long Short-Term Memory Recurrent Neural Network Architectures, and Random Forest have been used to predict Creutzfeldt-Jakob Disease Levels in the United States. New variables were created as data for the models to use on the basis of common factors that are known to affect CJD, such as soil, food, and water quality. Based on the root mean square error (RMSE), mean bias error (MBE), and mean absolute error (MAE) values, the study reveals the high impact of unhealthy lifestyle choices, $CO_2$ Levels, Pesticide Usage, and Potash $K_2O$ Usage on CJD Levels. In doing so, the study highlights new avenues of research for CJD prevention and detection, as well as potential causes.**

## 1 Introduction

First discovered between 1921 and 1923 by Hans Gerhard Creutzfeldt and Alfons Maria Jakob, Creutzfeldt-Jakob disease (CJD), commonly known as mad cow disease, is a rapidly progressive and fatal prion disease caused by an abnormal prion in the brain, termed PrP$^{Sc}$. PrP$^{C}$ is the harmless version of this prion and is found in abundance throughout the body and the nervous system. When ingested, PrP$^{Sc}$, that is often acquired from a non-native environment, attaches to the PrP$^{C}$ prion and causes a portion of the $\alpha$-helical and the coil structure of the PrP$^{C}$ protein to refold into $\beta$-sheets, converting the PrP$^{C}$ into its misfolded pathogenic isoform, PrP$^{Sc}$. Additionally, the misfolding encourages a resistance to proteases in the C-terminal region, resulting in the body's inability to destroy and remove PrP$^{Sc}$ from the brain. This structural change explains many physicochemical properties of PrP$^{Sc}$, such as their fibrillar structures, which reduces their susceptibility to degradation [3].

As a result, in its spread throughout the brain, the build-up and accumulation of PrP$^{Sc}$ results in lesions, damage to cells, gliosis, and neuronal loss. In specific, the accumulation of amyloid plaques, a common lesion caused by CJD, prompts neural and glial cell death and spongiform degeneration in the hippocampus, neocortex, basal ganglia, and thalamus. These areas of the brain control memory function, proper muscle coordination, higher-order thought processes, and transmission of information, respectively. Hence, CJD results in degradation in various areas of the brain, ultimately causing detrimental physiological effects that result in its multifaceted clinical presentation. The characteristics of the pathology often present increasingly worse symptoms ranging from depression, disorientation, motor difficulty, hallucinations, and can eventually develop into dementia or Alzheimer's. However, CJD has notably lacked any hallmark features due to its similarities to other neurodegenerative diseases. Unfortunately, this has hindered clinicians and scientists from determining a definitive diagnosis outside of a postmortem autopsy that would allow for treatment, making it all the more necessary to understand major factors that can implicate CJD [3].

In specific, physicians only understand the causation of hereditary CJD and variant CJD (vCJD), which account for 15% of CJD cases annually. In the hereditary form, a genetic mutation in the *PRNP* prion gene on chromosome 20 brings rise to PrP$^{Sc}$. The mutation is then passed down via autosomal dominant inheritance. In vCJD, the consumption of cattle products with an agent of bovine spongiform encephalopathy (BSE), is the underlying cause of a buildup of infectivity in the the lymphoreticular tissues and CNS. Furthermore, vCJD is linked to a methionine homozygosity at codon 129 in the *PRNP* prion gene, as the PrP$^{Sc}$ variant of vCJD is closely related to the codon 129 genotype, whose methionine homozygosity has been known to assist in prion strain propogation, allowing PrP$^{C}$ to be converted to PrP<sup>Sc</sup> [3].

[3] has been able to build off of this understanding, by establishing a notably strong correlation between CJD levels and beef production in the United States. Bhakta's findings reported an $r$ of 0.7632. Given that the results of the Pearson Correlation test provided an $r$ greater than 0.7, it indicates a strong potential that beef production can accurately predict CJD levels. However, to date, this is the only study that delves into this topic. Given the strong implications of Bhakta's work, we believe that with the proper combination of variables and a more diverse data set, a greater predictive power can be achieved.

In this work, 3 techniques – Elastic Net Regression (ENR), Long Short-Term Memory (LSTM) Recurrent Neural Network (RNN) Architectures, and Random Forest (RF) – have been used for prediction. Prior work with these techniques i.e. Stock Price Prediction, have shown very promising results in terms predictive power [9, 25]. The models introduce new variables, created via common factors that are known to affect ones susceptibility to prion and neurodegenerative diseases. With the introduction of these new variables, we believe that the we will achieve a higher degree of accuracy compared to those of prior works [3]. The accuracy of our models will be tested by 3 methods: RMSE, MAE, and MBE.

We found that without proper knowledge of disease contraction and pathophysiology, there is an inability to properly diagnose and effectively treat this fatal condition. Our study aims to broaden our understanding of CJD to allow for correct diagnosis and treatment. With the information we have compiled in this study, we believe we will be successfully able to determine which environmental and lifestyle factors can contribute to contracting CJD. Ultimately, we predict these findings will aid physicians and researchers to determine specific causes of CJD and identify symptoms of CJD that lead to early diagnosis and treatment.

The rest of this paper is organized as follows. Section 2 discusses the data set and pre-processing properties of this study, such as new variables and training and testing sizes. Section 3 provides preliminaries to deep learning and machine learning, before explaining the ENR, LSTM, and RF models that we use in this study. Section 4 gives the results of each model as well as compares the accuracy of each model based on the methods highlighted above, with section 5 concluding the paper and talking about future opportunities that the work provides.

## 2  Data Set and Data Pre-Processing

In this section, we will discuss the data set used in this study, new variables that were introduced, and intrinsic properties of the data set.

### 2.1  Data Set and New Variables

In this work, we introduce eight new variables to predict CJD levels in the United states from 1979 to 2015, that along with beef production levels from Bhakta's work, form our data set. The eight new variables have been created to aid in more accurate prediction of CJD levels, based on common factors that are known the affect similar prion and neurodegenerative diseases, such as the environmental contamination of soil, food, and water [6]. The new variables are:

1. $CO_2$ Levels (Yearly) [21]
2. Nitrogen Usage (Yearly) [12]
3. Potash $K_2O$ Usage (Yearly) [12]
4. Pesticide Usage (Yearly) [11]
5. Beer Consumption (Yearly) [15]
6. Obesity Levels (Yearly) [14]
7. Smoking Levels (Yearly)[13]
8. Smoking Less Than 1 Pack (Yearly) [13]

Sources for each variable are included next to them. Data for $CO_2$ Levels was taken from the National Oceanic and Atmospheric Administration's Global Monitoring Laboratory. Data for Beer Consumption was taken from the National Institute on Alcohol Abuse and Alcoholism. Data for Obesity Levels was taken from the National Center for Health Statistics-Centers for Disease Control and Prevention. Data for Smoking Levels and Information was taken from the Gallup's Tobacco and Smoking Database. Data for Pesticide Usage, Potash $K_2O$ Usage, and Nitrogen Usage were taken from the Food and Agriculture Organization of the United States's Database.

Given that habits such as smoking and drinking, and an overall unhealthy lifestyle can cause one to be susceptible to diseases like Alzheimer's and Dementia, which share very common characteristics to CJD, we believe that introducing such variables to our model, will greatly improve predictive power. Additionally, as Pesticide, Nitrogen, Potash $K_2O$, and $CO_2$ usage can all contribute to soil and water contamination, we feel that introducing such variables will greatly improve accuracy and reveal possible sources of the currently unknown causes of CJD.

Yearly CJD case levels were taken from the Centers for Disease Control and Prevention's (CDC) Creutzfeldt-Jakob Disease, Classic (CJD) database [5]. All of the data was normalized using Standard Scalar fit to ensure consistency in the data set and that each data type has the same format. Information about the training and testing set are shown in Table 1. The training data is used to construct the model, while the testing data tests the validity of the trained model.

### 2.2  Intrinsic Data Set Properties

Looking at the variables that we propose, it is important to understand the relationship between them and the potential impact this has on the study. We first consider the

Table 1: Relationship between variables on CJD Levels (rounded to 3 decimals)

| Variables | $r$ | $R^2$ | $t-statistic$ | $\alpha$ | $p$ |
|---|---|---|---|---|---|
| Beef Production | -0.824** | 0.679 | -8.489¶ | 0.904†† | 6.478×10⁻¹⁰ ‡ |
| Beer Consumption | -0.879** | 0.773 | -10.746¶ ¶ | 0.936†† | 1.789×10⁻¹² ‡ |
| $CO_2$ Levels | 0.909*** | 0.826 | 12.735¶ ¶ | 0.953†† | 1.706×10⁻¹⁴ ‡ |
| Nitrogen Usage | -0.808** | 0.653 | -8.000¶ | 0.894† | 2.535×10⁻⁹ ‡ |
| Obesity Levels | 0.837** | 0.701 | 8.924¶ | 0.911†† | 1.979×10⁻¹⁰ ‡ |
| Pesticide Usage | 0.916*** | 0.840 | 13.356¶ ¶ | 0.956†† | 4.394×10⁻¹⁵ ‡ |
| Potash $K_2O$ Usage | -0.753* | 0.567 | -6.470¶ | 0.859† | 2.796×10⁻⁷ ‡ |
| Smoking Less than 1 Pack | 0.849** | 0.721 | 9.376¶ | 0.918†† | 5.908×10⁻¹¹ ‡ |
| Smoking Levels | -0.825** | 0.681 | -8.512¶ | 0.904†† | 6.086×10⁻¹⁰ ‡ |

*$r \geq 0.7$   **$r \geq 0.8$   ***$r \geq 0.9$   ‡$p \leq 0.001$   †$\alpha \geq 0.85$   ††$\alpha \geq 0.9$
¶ $t-statistic \geq 5 \cup t-statistic \leq -5$   ¶¶ $t-statistic \geq 5 \cup t-statistic \leq -5$

Table 2: Statistics for the Data Set

| Data Set | Training Data Set | Testing Data Set |
|---|---|---|
| 1979-2015 | 70% | 30% |

relationship between the proposed variables. Figures 1 and 2 depict the correlation between the proposed variables. As seen, although all the variables used in this study are unique, there are high degrees of correlation between them ($r \geq 0.5$). Current Recurrent Neural Networks and Random Forest techniques are not affected by multicollinearity , however, standard MSE multivariate regression models are greatly impacted by multicollinearity, due to its ability to reduce precision in estimating coefficients and ergo, weaken the statistical significance of the model. Thus, in this paper, we will not use a standard MSE multivariate regression, but will instead favor an ENR model, which performs favorably with multicollinearity, as we will see in section 3.2.

Next, we investigate the correlation between proposed variables and CJD levels, to ensure their use will yield accurate and significant results. We determined this using Pearson Correlation and Cronbach's Alpha. The findings for these tests are summarized in Figure 2 and Table 2.

All variables exhibited an $r$ value greater than 0.7, with 22% yielding an $r$ value greater than 0.9. Additionally, all variables had a $p$ value less than 0.001 (less than the minimum value of 0,05), indicating that our results were significant. Such high $r$ values lead us to believe that our proposed variables may have potential in providing an accurate prediction of CJD levels, but also highlight that individually all of these variables serve as potential markers for CJD prediction (as we will explore more in section 4). Moreover, $\alpha$ values greater than 0.85 and $t-statistic$ values greater than 5, reveal high degrees of internal consistency and low levels of variance, which emphasizes that the data is accurate and consistent over the span of all columns and that all of our proposed variables have potential as strong predictors of CJD levels.

## 3 Preliminaries

In this section, we first provide background information about deep learning and neural networks, specifically, RNNs. Then we give an overview of Elastic Net Regression, Long Short-Term Memory Recurrent Neural Network Architectures, and Random Forest.

### 3.1 Deep Learning and Neural Networks

RNNs are a subset of Deep Learning that have found tremendous accuracy in pattern recognition and feature extraction. Conventional neural networks, such as Convolutional Neural Networks (CNNs) and Deep Neural Networks (DNNs) have been known to be incapable of using prior outputs to better process and understand inputs in the current step, due to their fixed-size vectors as inputs and fixed number of computational steps. However, RNNs bypass this restriction with a unique operation over a sequence of vectors over time that uses outputs from the output units as inputs of the hidden layer [2]. Formally, this can be represented via the following equation:

$$h_t = \sigma_h(w_h x_t + u_h y_{t-1} + b_h)$$
$$y_t = \sigma_y(w_y h_t + b_y) \quad (1)$$

Where $x_t$ is a vector of inputs, $h_t$ are hidden layer vectors, $y_t$ are output vectors, $w$ and $u$ are weight matrices, and $b$ is the bias vector.

Hence, the loop utilized by RNNs that passes information from one step of the neural network to the next, allows for greater pattern recognition and predictive modeling, which will permit us identify the most significant environmental and lifestyle habits that contribute to CJD infection.

### 3.2 Elastic Net Regression

In traditional regression modeling, variable selection and prediction from high dimensional data sets can be problematic with low sample sizes, as is the case in this paper, due to the rarity of CJD. However, Elastic Net Regression generalizes many shrinkage-type regression methods, such as Ridge and Lasso regression, to bypass this issue. In specific, using the least angle regression algorithm, it estimates coefficients in a biased manner and improves accuracy by shrinking the estimated parameters, to reduce variance [10]. This is done though the tuning of the penalty term:

$$(\frac{1}{2}(1-\alpha)\beta^2 + \alpha|\beta|) \quad (2)$$

through the parameter $\alpha$, which has been generalized into the following equation.

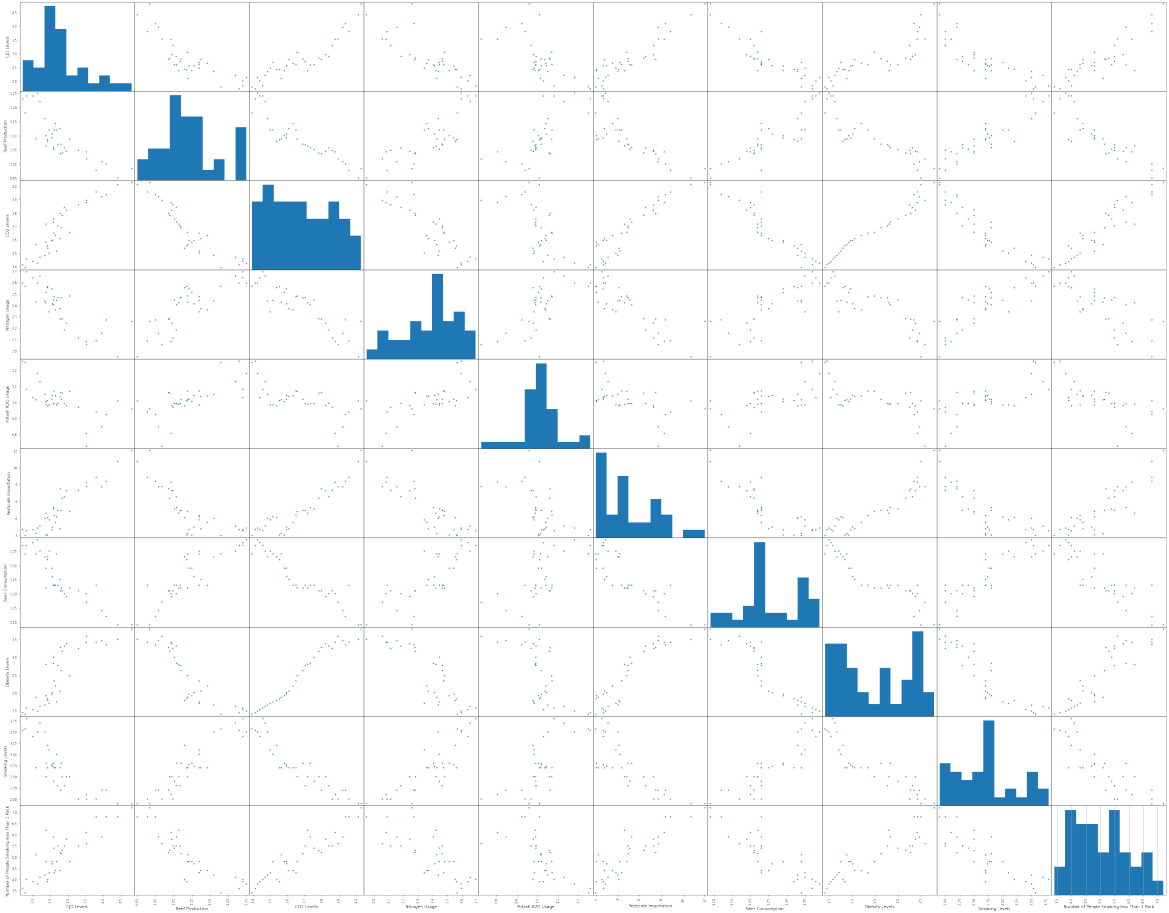

Figure 1: Scatter plots between newly proposed variables and CJD Levels on each other

$$\hat{\beta} = \underset{\beta}{\mathrm{argmin}} \left\{ \sum_{i=1}^{N} \left( y_i - \sum_{j=1}^{P} x_{ij}\beta_j \right)^2 + \lambda_1 \sum_{j=1}^{P} |\beta_j| + \lambda_2 \sum_{j=1}^{P} \beta_j^2 \right\} \quad (3)$$

Where $\alpha$ controls the type of shrinkage, and the penalty parameter $\lambda$ controls the amount of shrinkage.

As previously stated, Elastic Net Regression is the combination of Lasso and Ridge regression. Lasso ($\alpha = 1$) has an $l_1$ penalty to control both parameter shrinking and variable selection, whereas Ridge ($\alpha = 0$) only has an $l_2$ penalty on the parameters. By combining both regression methods, $\alpha$ can be $0 < \alpha < 1$, Elastic Net Regression allows for the integration of different weights of automatic variable selection and no variable selection into the model. Furthermore, in utilizing both Lasso and Ridge regression, Elastic Net Regression is able to overcome the traditional problem of of correlated predictors, by subtracting a small number $\epsilon$ from $\alpha$, so that in its parameter shrinkage, more correlated predictors can be incorporated in the model. Essentially, Elastic Net Regression allows for a balance between penalties by amalgamating feature elimination and feature coefficient reduction to enable a model to have more effective and powerful prediction [10].

An Elastic Net Regression model's ability to properly handle multicollinearity, suits our work's needs, due to the high levels of multicollinearity in the model (see section 2.3). Additionally, we believe that its integration of different weights for variables will allow us to understand which variables hold the most weight in CJD prediction.

### 3.3 Long Short-Term Memory

As mentioned in section 3.1, RNNs allow for greater pattern recognition and predictive modeling, when compared to more traditional neural networks; however, a recurring problem with RNNs has been the vanishing gradient problem. The problem stems from the fact that as more layers are added to neural networks with certain activation functions, the gradient of the loss function approaches zeros, contributing to a larger degree of inaccuracy in a model. In looking at certain activation functions, a large input is scaled into a small input space between 0 and 1. By doing so, although a large change may occur in the input, there will be a small change in the output, leading to the derivative becoming smaller than expected. As a result, since the gradients of neural networks are derived via backpropogation, due to chain rule, as an increased amount of layers are added to the network and the derivatives are multiplied from the final layer to the initial to calculate the derivative of the initial layers, there will be an exponential decrease in the gradient that directly correlates with the propagation to the initial layers:

$$\frac{\partial \varepsilon}{\partial \theta} = \sum_{1 \leq t \leq T} \frac{\partial \varepsilon_t}{\partial \theta}$$
$$\frac{\partial \varepsilon_t}{\partial \theta} = \sum_{1 \leq k \leq t} \frac{\partial \varepsilon_t}{\partial x_t} \frac{\partial x_t}{\partial x_k} \frac{\partial^+ x_k}{\partial \theta} \quad (4)$$
$$\frac{\partial x_t}{\partial x_k} = \prod_{t \geq i \geq k} \frac{\partial x_i}{\partial x_{i-1}} = \prod_{t \geq i \geq k} W_{rec}^T diag(\omega'(x_{i-1}))$$

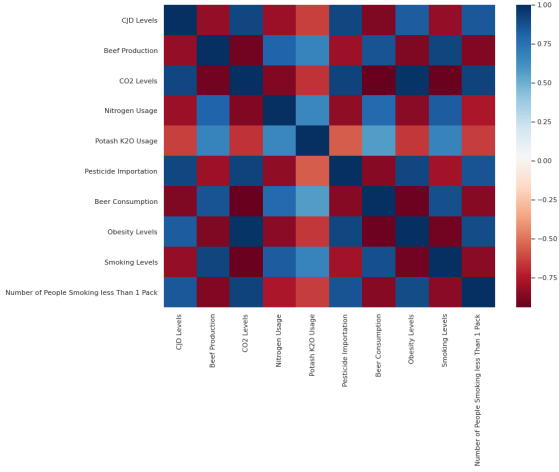

Figure 2: Correlation between variables in the model

Where $\varepsilon$ is the different outputs of the model, $x$ is the the different inputs, and $W_{rec}$ represents the backpropagation algorithm [16].

Consequently, the weights and biases will not be updated correctly with each training layer, leading to a much lower degree of accuracy [16]. Nonetheless, LSTM models look to overcome the vanishing gradient problem by allowing for a constant error flow though self-connected units that are capable of learning long-term dependencies, and in turn, avoid them. As opposed to the standard RNN that has a very simple chain of repeating modules, LSTMs introduce a much more chain-like repeating module that consists of 4 layers [17].

The main purpose of doing so, is to maintain the cell state of the model, or a linear set of information flowing from cells states, $C_{t-1}$ to $C_t$. This state is heavily regulated and is only affected by minor linear interactions at 3 gates composed of a sigmoid neural net layer and a point wise multiplication operation. In the first step, a sigmoid layer, referred to as the "forget gate layer" looks at the inputs from $h_{t-1}$ and $x_t$ and outputs a value between 0 and 1 for each number in the $C_{t-1}$ state, that indicates whether to keep certain inputs in the model. Then, a sigmoid layer, known as the "input gate layer" determines which values in the model will be updated and a $tanh$ layer produces a vector, $\tilde{C}_t$, for new candidate values. Next, the model multiplies the $C_{t-1}$ state by $f_t$ to forget the values that had already been determined as unnecessary, before adding the new canidate values, $i_t \times \tilde{C}_t$, to update the cell state to $C_t$. Finally, using a sigmoid layer to decide which components of the cell state to output, the model then runs the cell state through $tanh$ and multiplies it by the output of the sigmoid gate to generate an output [17]. This multi-step LSTM process can be represented with the following equation:

$$\begin{aligned} f_t &= \sigma(W_f \times [h_{t-1}, x_t] + b_f) \\ i_t &= \sigma(W_i \times [h_{t-1}, x_t] + b_i) \\ \tilde{C}_t &= \tanh(W_C \times [h_{t-1}, x_t] + b_C) \\ \sigma_t &= \sigma(W_o \times [h_{t-1}, x_t] + b_o) \\ h_t &= o_t \times \tanh C_t \end{aligned} \quad (5)$$

Where $C_{t-1}$ and $C$ are the old and new cell states, respectively, $h_{t-1}$ and $x_t$ are the inputs, $\sigma$ is the sigmoid layer, $f_t$ is the input to the first gate, $i_t$ and $\tilde{C}_t$ and $i_t$ are the inputs to the second gate, $\sigma_t$ is the input to the third gate, and $h_t$ is the output.

This study makes use of this improvement, by creating 3 layers in our model: an input layer, hidden layer, and output layer. The input layer consists of our newly proposed variables, as stated in section 2.2. The weights on each input are calculated and then sent to the hidden layer. At each gate, the total weight is derived and sent to the output layer, which consists of one neuron, being the predicted value. By overcoming the vanishing gradient problem, we anticipate that it will allow us to observe the high levels of accuracy that our newly proposed variables provide.

### 3.4 Random Forest

Random Forest is an ensemble of decision tree predictors that is capable of regression and classification tasks. The main idea behind the ensemble approach is that multiple individual classifiers can merge together to form a much stronger classifier and more accurate and stable prediction. However, rather than rely on individual decision trees, in which an input traverses down the tree into smaller sets to determine the output, Random Forest reduces variance in the model by training multiple decision trees in parallel with bootstrapping proceeded by aggregation, which is known as bagging. In specific, the bootstrapping allows for the individual decision trees to be trained using different subsets of available features in the data set while also guaranteeing that each such decision tree is unique of every other [22].

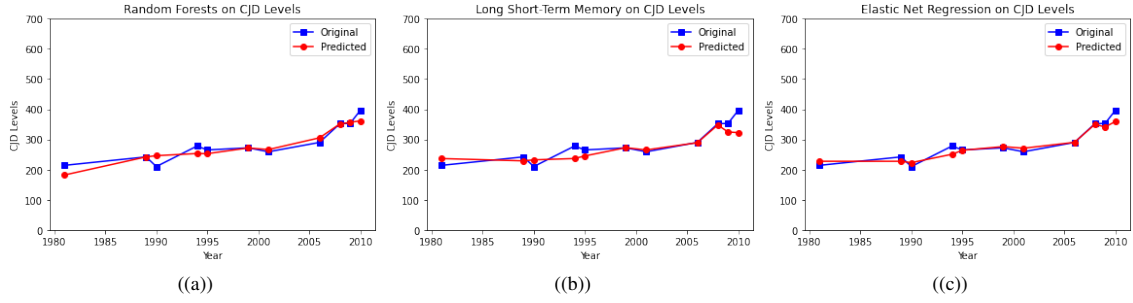

Figure 3: CJD Level Predicted with (a) Random Forest (b) Long Short-Term Memory Recurrent Neural Network Architecture (b) Elastic Net Regression

It does this by taking a random subset with replacement from the input, and then taking another random subset of predictor variables at each node. Then, using the predictor variable that gives the best split, it performs a binary split at the node. In doing so, as each new input enters the model, it is put through all of the trees to output a weighted average of all the terminal nodes, or when there are categorical variables, a voting majority. Accordingly, by aggregating the outputs from each individual decision tree, the Random Forest classifier ensures a good generalization. In addition, because of its vigorous selection process for training samples, Random Forest classifiers reduce noise in the training data set, and in turn, are able to increase the overall accuracy without over fitting the model. [22].

In this work, our proposed variables are provided for training in each tree, in order to determine each decision at each node. Further, due to the lack of prior information about CJD, the trees reduce error by approaching the analysis as a classification problem and basing forecasts on prior training variables. In doing so, we believe that the Random Forest will reveal the true extent to which select environmental and lifestyle choices affect ones susceptibility to CJD.

## 4 Results

In this section, we study the effects of introducing our new proposed variables on the Random Forest model, the Long Short-Term Memory Recurrent Neural Network Architecture, and the Elastic Net Regression model.

To evaluate the effect that our newly proposed variables have on the precision of the model in predicting CJD levels, we use 3 methods to measure error. These methods include Root Mean Square Error (RMSE), Mean Bias Error (MBE), and Mean Absolute Error (MAE).

RMSE is calculated using the following equation:

$$RMSE = \sqrt{\frac{\sum_{i=1}^{n}(O_i - F_i)^2}{n}} \qquad (6)$$

Where $O_i$ refers to the actual CJD levels, $F_i$ refers to the predicted CJD levels, and $n$ refers to the total window size. In this paper, we define an RMSE value of less than 1 as statistically significant.

MBE is computed using:

$$MBE = \frac{1}{n}\sum_{i=1}^{n}(O_i - F_i) \qquad (7)$$

Where $O_i$ is the actual CJD level, $F_i$ is the predicted CJD level, and $n$ refers to the total window size. We define take an MBE value of less than 5 as statistically significant.

MAE has also been used to performance of the models, via the following equation:

$$MAE = \frac{1}{n}\sum_{i=1}^{n}|O_i - F_i| \qquad (8)$$

Where $O_i$ refers to the actual CJD levels, $F_i$ refers to the predicted CJD levels, and $n$ refers to the total window size. In this paper, we define an MAE value of less than 5 as statistically significant.

Figure 3 (a) represents the original CJD levels with respect to the predicted CJD levels using RF. Figure 3(b) represents the original CJD levels with respect to the predicted CJD levels using LSTM. Figure 3 (c) represents the original CJD levels with respect to the predicted CJD levels using ENR. In building our ENR model, we found that an $\alpha$ value of 0.0017 provided for optimal predictions, as seen in Figure 4. Comparative analysis of RMSE, MBE, and MAE values obtained using RF, LSTM, and ENR is shown in Table 2. It can be observed that ENR performs with the most accuracy, as seen by having the lowest RMSE and MAE values of 0.179 and 0.136 respectively.

Given that the ENR model provided the most accurate results via the comparative analysis, we will be utilizing its variable weightings to see which of the newly introduced variables were weighted with the most significance, and hence, have the greatest impact on predicting contraction of CJD. The absolute value of each coeffient represents its weight in terms of predicting CJD. Larger values therefore represent factors that increase risk of CJD contraction, while lower values represent lower risk of CJD contraction. The weightings obtained by the ENR model are shown in Table 4.

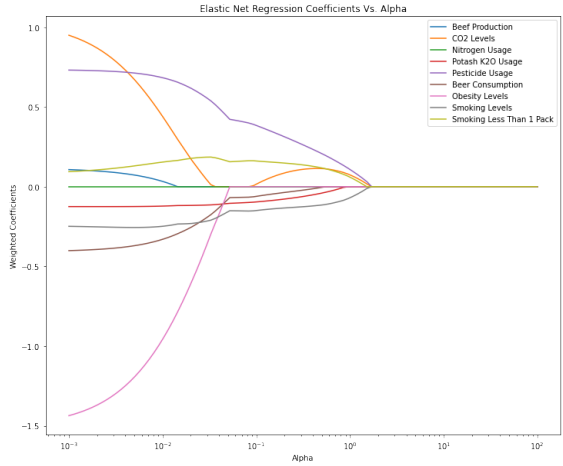

Figure 4: Alpha Value for the Elastic Net Regression Model

Table 3: Comparative analysis of RMSE, MBE, and MAE values obtained by ENR, LSTM, an RF Models (rounded to 3 decimals)

| Model Type | RMSE | MBE | MAE |
|---|---|---|---|
| RF | 0.204 | 0.041 | 0.154 |
| LSTM | 0.297 | 0.118 | 0.224 |
| ENR | 0.179 | 0.046 | 0.136 |

Table 4: Coefficients of the Elastic Net Regression Model (rounded to 3 decimals)

| Variable | Coefficient |
|---|---|
| Beef Production | 0.101 |
| $CO_2$ Levels | 0.895 |
| Nitrogen Usage | 0.005 |
| Potash $K_2O$ Usage | -0.125 |
| Pesticide Usage | 0.729 |
| Beer Consumption | -0.396 |
| Obesity Levels | -1.391 |
| Smoking Levels | -0.251 |
| Smoking Less Than 1 Pack | 0.102 |

The weightings of the ENR model indicate that lifestyle choices, such as Beer Consumption (-0.396), Obesity (-1.391), and Tobacco Usage (-0.251), can lead to a host of pathologies - ultimately leading to a weakened immune system and increasing risk of contracting CJD. Additionally, the coefficients provided that pollution and $CO_2$ Levels (0.895) also greatly impact CJD levels. However, the weightings also show that factors such as Pesticide Usage (0.729) and Potash $K_2O$ Usage (-0.125) also affect ones susceptibility to contracting CJD. The implications of these weightings are discussed in Section 5. Variables such as Beef Production (0.101) and Nitrogen Usage (0.005) had the lowest weightings, suggesting that they may not have a large impact on whether one acquires CJD. Moreover, the weighting for the Beef Production variable is consistent with the known percentage of vCJD cases caused by BSE each year (10% to 15% each year). As BSE and gene transfer are the only known causes for CJD (vCJD and hereditary CJD, respectively) a proper weighting for the Beef Production variable suggests that in addition to having highly significant RMSE, MBE, and MAE values, the models are also in accordance with the current understanding for the causes of CJD.

## 5 Discussion

In this section, we discuss the importance of our findings, their implications, and how they relate to lifestyle factors, environmental factors, and agricultural production. Our findings elucidate several factors which predict increased chance of contracting CJD. Our hope is that these results can provide evidence for regulating agricultural products, and providing information for the general public regarding the impacts of improper diet and poor lifestyle habits.

Our results provided that Beer Consumption, Obesity, and Tobacco Usage all have a great impact on CJD levels. It is known that a person suffering from chronic obesity is at higher risk of developing diseases such as diabetes, high cholesterol and high blood pressure. Likewise, regular smokers are often susceptible lung disease and heart disease. Moreover, alcoholism, as seen in regular beer consumption, has been known to chronically damage the stomach lining, digestive tract, and liver leading to a host of chronic diseases such as fatty liver or gastritis. Chronic alcohol consumption has been known to inhibit the body's ability to extract important nutrients from food, therefore contributing to more disease in the long run. These pathological conditions enable an infectious host, such as CJD, to utilize the host's susceptibility to disease to its advantage and successfully invade healthy brain tissue. Given that the majority of energy is being expended to correct these already existing pathologies, any co-morbid infections are more likely to cause disease as the immune system is already overwhelmed. As the body already contains a high degree of $PrP^c$, we believe that the body's inability to fight off the pathological variant of CJD is due to $PrP^{Sc}$'s similarity to the physiological prion that is common in the CNS; the immune system therefore does not recognize that it has a pathophysiological condition invading and thus, does not respond in the appropriate manner, providing an explanation for why the adaptive immune system does not respond to CJD.

Though $CO_2$ is a natural component of our body's metabolic processes, rising levels of environmental $CO_2$ have contributed to rising CJD levels. Over the past 40 years, $CO_2$ levels have increased 20% from 339 part per million to 410 parts per million [20]. In saying so, an increased concentration of atmospheric $CO_2$ causes increased levels in urban and indoor levels, which can lead to its build up in poorly ventilated spaces over long periods of time. When this occurs, one is more likely to breathe air with high $CO_2$ levels, causing an increased amount of $CO_2$ to be in their blood stream. This in turn reduces the amount of $O_2$ that reaches one's brain and can cause significant cognitive impairment, such as difficulties with decision-making and planning, all of which are common in CJD cases [19]. Those whose vocation involves working in industrial or secluded areas are more likely to contract CJD due to decreased ventilation and air supply, and hence a higher chance for an increased concentration of $CO_2$ to be in the air [7]. We believe that this lends itself to the proposition that a higher probability of an increased amount of $CO_2$ in the bloodstream of those with CJD can lead to their susceptibility to the disease. This is because the immune system depends on $O_2$ in the blood as a regulator of immune responses and energy consumption of immune effectors, by controlling the immunoregulatory activity in the body [23]. Hence, a decreased amount of $O_2$ being delivered to the brain, may lead to an under performing immune system in neural tissue and the immune system's inability to properly detect and react to the pathophysiological condition, much like for the unhealthy lifestyle factors we have identified.

In identifying potential areas that contribute to the immune system's inability to properly detect and react to a CJD infection, one key question remains: what leads to CJD's neurodegenerative properties? Our results provide a potential contributor to these characteristics. It has previously been identified that CJD originates from a source outside of the body and can be caused by contamination of soil, food, and water [3, 6]. Our study identified Pesticide Usage and Potash $K_2O$ Usage as key contributors to predicting CJD levels and potential contributors to CJD infection. Given the extremely high weighting of the Pesticide Usage variable (0.729) in the ENR model, and the relatively moderation weighting of the Potash $K_2O$ Usage variable, we believe that there is a likelihood that features of both substances contribute to CJD's neurodegenerative properties. Prior studies have highlighted how acute high-levels of exposure to pesticides have neurotoxic effects, can cause decreases in neurobehavioral performance, reflected in cognitive and psychomotor dysfunction [18]. However, no previous studies have linked exposure to pesticides to the characteristic spongiform neurodegeneration of CJD. We suspect that it can be linked to 3 main common compounds that are present in pesticides: organochlorine compounds, organophosphate, and neonicotinoids.

Organochlorine compounds, also known as chlorinated hydrocarbons, work by opening the sodium ion channels of neurons and causing them to fire spontaneously and degenerate. As pesticides are designed to kill insects, they often cause them to spasm and eventually die.

Likewise, organophosphates also work on the nervous system, but they prevent nerve cells from communicating with each other. Under normal conditions, nerve cells in the brain send electrical pulses down the tendril, where the pulse jumps across the synapse to another nerve cell. Additionally, a chemical compound know as ACh moves from one to the other and binds with the adjacent neuron, transferring the electrical impulses. This propagation of signals allows for communication amongst neurons. However, organophosphates prevent the transfer of ACh and limit the activity in certain parts of the brain. These effects can impair communication so significantly that widespread paralysis can be seen in those infected with large amounts of organophosphates.

Similarly, neonicotinoids are the synthetic version of nicotine. They affect nerve cells by strongly binding to nicotinic acetylcholine receptors in the central nervous system, causing overstimulation of the nerve cells, and eventual disorientation, paralysis, and death

Much like drugs derived from plants that instantaneously kill animals and insects, we hypothesize that these common chemicals have a more longitudinal impact on humans. Rather than cause an immediate neurodegeneration, a continual consumption of chlorinated hydrocarbons, organophosphates, and neonicotinoids, whether in food or in an agricultural setting, can accumulate to lead to long-term pathologies. As highlighted above, organochlorines and neonicotinoids overstimulate the sodium ion channels and nicotinic acetylcholine receptors, respectively, to cause a slow disorientation, paralysis, and death that is characteristic of CJD. Additionally, we believe that the ultimate spread of the $PrP^{Sc}$ prion throughout the brain is caused in part by organophosphates. Rather than being able to signal to the rest of the nervous system that it is degenerating, organophosphates avert electrical impulses of nerve cells and hence reduces the reactivity of the immune system. In doing so, as well as unhealthy lifestyle choices and environmental changes that divert the immune system's response, the nerve cells' inability to signal their distress allows the $PrP^{Sc}$ prion to cause spongiform degeneration throughout the nervous system, without an activation of the adaptive immune system.

Moreover, our results provided that Potash $K_2O$ has potential as a contributor to CJD. Although little is known about the effect of Potash $K_2O$ on the brain, we speculate that it acts through 2 main mechanisms. First, much like how organochlorines and neonicotinoids overstimulate the sodium ion channels and nicotinic acetylcholine receptors, respectively, an increased consumption of compounds of Potash $K_2O$ could cause the potassium sodium pumps in the nervous system to shoot irregularly. The body contains many oxidation chains, through which $K_2O$ can be converted to a potassium (K) containing output. As a result, the K concentration in the plasma can cause an irregular stimulation of the the potassium sodium channels, which are integral for information processing and cell-to-cell communication in the nervous system. Hence, similar to the proposal of the impact that an increased concentration of organophosphates compounds has on the body, an increased consumption of Potash $K_2O$ can skew the body's immune response.

Second, we believe that Potash $K_2O$ contributes to CJD's neurodegenerative properties via its inhibition of the activity of aldehyde dehydrogenase (ALDH) in the brain. ALDH detoxifies 3,4-dihydroxyphenylacetaldehyde (DOPAL), the oxidized form of the neurotransmitter dopamine. Dopamine is vital for a variety of biological functions, including motor function. However, when it oxidizes via monoamine oxidase, dopamine transforms into a highly toxic compound, known as DOPAL. ALDH enzymes work to reverse the toxic production of DOPAL by carbonyl metabolism and convert it into a less toxic acid product. An inhibition of ALDH though, causes an accumulation of DOPAL in the brain that results in an alteration to dopaminergic cells, by modifying proteins and enabling protein aggregation [8]. As cellular prion proteins are present in dopaminergic neurons and modulate the dopaminergic system, we theorize that in combination with high levels of pesticides in the surrounding envionment, substantial levels of Potash $K_2O$ could also be contributing to the neurodegenerative properties of CJD [24]. As mentioned above, in being oxidized, $K_2O$ is converted to K. It has previously been discerned that K can inhibit ALDH, while promoting dopamine production [4, 1]. In turn, Potash $K_2O$ is enabling the spread and build up of DOPAL throughout the brain, which with its toxic properties is causing the degeneration of prion proteins, that accumulate into amyloid plaques. As a result, Potash $K_2O$ aids in neurodegeneration and allows for the formation of characteristic lesions of CJD.

## 6 Conclusion

Identifying unique biomarkers and characteristic causes of CJD is a challenging task due to the historic lack of information around the disease. Prior studies have highlighted BSE as a cause for vCJD and genetic mutations on the *PRNP* prion gene as a cause for hereditary CJD. However, these findings only account for 10% to 15% of CJD cases annually.

To obtain a more in-depth understanding of where the $PrP^{Sc}$ prion's neurodegenerative qualities stem from, we introduced 8 new variables to a data set with existing variables (Beef Production). These new variables are categorized as factors that cause environmental contamination of soil, food, and water, which has previously been identified as causes of similar prion and neurodegenerative diseases. Using the new data set as testing and training sets, RF, LSTM, and ENR are used to to predict historic CJD levels. The comparative analysis based on RMSE, MBE, and MAE values clearly indicated that the ENR model gave the best prediction of CJD levels. Results showed that the ENR model gives RMSE (0.179), MBE (0.046), and MAE (0.136), which we defined as being extremely accurate and significant comparative analysis values.

In observing the accuracy of the ENR model, its weighting of variables was further investigated. These weightings revealed that unhealthy lifestyle habits, such as obesity, smoking, and beer consumption, along with rapid greenhouse gas emission and pollution contribute to CJD infection. Nonetheless, it was also revealed pesticides and fertilizers, in specific, their chlorinated hydrocarbons, organophosphates, neonicotinoids, and potash $K_2O$ compounds, as potential contributors to the characteristic spongiform degeneration of CJD. In specific, chlorinated hydrocarbons and neonicotinoids were attributed to nerve cell death, while organophosphates and potash $K_2O$ compounds were found to mitigate the nerve cell communication and activation of the adaptive immune system. Furthermore, it was proposed that potash $K_2O$ contributed to the inhibition of ALDH in the brain, which likewise aided in neurodegeneration but ultimately may be an underlying reason for the accumulation of $PrP^{Sc}$ and the formation of amyloid plaques.

For future work, further clinical studies can be performed for those in the presence of high concentrations of chlorinated hydrocarbons, organophosphates, neonicotinoids, and potash $K_2O$ compounds. In particular it will be important to investigate how those in an agricultural setting are more vulnerable to CJD infection. Furthermore, deep learning models could be developed, which using the identified variables, could provide an opportunity for early diagnosis of CJD. Upon diagnosis, developing case studies to understand the disease model of CJD could be pertinent in contributing to treating symptoms and delaying disease disease progression.

## 7 Acknowledgments

The authors are grateful for the generous support of Dr. Jithendra Kini Bailur of Takeda Oncology and the Department of Medicine at the Yale University School of Medicine, for his critical work in reviewing the methodologies of this paper.

Address correspondence to: Arnav Bhakta, Phillips Academy Andover, 180 Main Street, Andover, Massachusetts, USA. Phone: 978.770.1575; Email: abhakta22@andover.edu.